\documentclass[letterpaper, 10 pt, conference]{ieeeconf}  

\IEEEoverridecommandlockouts                              

\usepackage{amsmath} 
\usepackage{amssymb}  
\usepackage{graphicx}

\usepackage{color}

\title{\LARGE \bf
	A Short Note on Analyzing Sequence Complexity in Trajectory Prediction Benchmarks	
}

\author{Ronny Hug$^{\dagger}$, Stefan Becker$^{\dagger}$, Wolfgang H\"ubner$^{\dagger}$ and Michael Arens$^{\dagger}$
	\thanks{$^{\dagger}$Fraunhofer IOSB, Ettlingen, Fraunhofer Institute of Optronics, System Technologies and Image Exploitation, 76275 Ettlingen, Germany
		{\tt\small firstname.lastname@iosb.fraunhofer.de}}%
	\thanks{Fraunhofer IOSB is a member of the Fraunhofer Center for Machine Learning.}%
}

\begin{document}  
	
\maketitle
\thispagestyle{empty}
\pagestyle{empty}

\begin{abstract}
	
The analysis and quantification of sequence complexity is an open problem frequently encountered when defining trajectory prediction benchmarks.
In order to enable a more informative assembly of a data basis, an approach for determining a dataset representation in terms of a small set of distinguishable prototypical sub-sequences is proposed.
The approach employs a spatial sequence alignment, which enables a following learning vector quantization (LVQ) stage.
A first proof of concept on synthetically generated and real-world datasets shows the viability of the approach.

\end{abstract}

\section{Introduction} 
An open question frequently encountered in the context of trajectory prediction is the quantification of dataset complexity, leading to hard-to-solve or too simple benchmarks.
Current attempts in standardized benchmarking, e.g. \emph{TrajNet++} as mentioned in \cite{rudenko2019human}, originate from heuristics or experience-based criteria when assembling the data basis.

Dissecting trajectory datasets, each dataset can be reduced to a small number of prototypical sub-sequences specifying distinct motion patterns, where each sample can be assumed to be a variation of these prototypes. 
Following this assumption, the quantification of dataset complexity could be based on the number and variation of prototypes.
Towards this end, an approach for determining prototypes from a dataset is proposed.
The approach applies a spatial alignment\footnote{Not to be confused with a temporal sequence alignment} step followed by vector quantization for clustering aligned samples. 
Experiments are conducted on a synthetically generated dataset and an exemplary standard benchmarking dataset in order to provide a first proof of concept.
Lastly, potential uses of the proposed approach are discussed.

\section{Sequence Alignment}
\label{sec:da}
Given a set of trajectories (\emph{samples}) $\mathcal{X} = \{X_1, ..., X_N\}$, as sequences of $M$ subsequent points $X_i = \{\mathbf{x}^i_1, ..., \mathbf{x}^i_M\}$, each sample is first normalized by moving it into a self-centered reference frame and scaling it to unit length:
\begin{align}
	X^{norm}_i = \left\{\frac{\mathbf{x}^i_j - \bar{\mathbf{x}}}{\mathbf{x}^i_M - \mathbf{x}^i_1} | j \in [1,..,M] \right\}.  
\end{align}
It has to be noted that this normalization solely serves the purpose of moving all samples into a common value range and is not a good normalization in terms of pooling similar samples. 
Then, all samples are aligned with a single prototype $\hat{Y} = \left\{\mathbf{\hat{y}}_1, \mathbf{\hat{y}}_2, ..., \mathbf{\hat{y}}_M \right\}$ by using similarity transformations, which are retrieved from a regression model $\phi: X \rightarrow \{\mathbf{t},\alpha,s\}$.
$\hat{Y}$ and $\phi$ are learned by minimizing the mean squared error between each aligned sample $X^\phi_i = \phi(X^{norm}_i)$ and the prototype $\hat{Y}$
\begin{align}
\label{eq:loss_da}
	\mathcal{L}_{align}(\phi(X^{norm}_i), \hat{Y}) = \frac{1}{M} \sum^M_{j=1} \|\mathbf{x}^i_j - \mathbf{\hat{y}}_j\|^2_2  
\end{align}
using stochastic gradient descent.
This is different from linear factor models, where the whole training set has to be considered.
With respect to equation \ref{eq:loss_da} and the similarity transformation, the trivial solution that maps all samples onto the origin has to be avoided.
A brute force approach to this problem is to enforce a minimum scale. 
These steps result in a minimum variance alignment of all samples with respect to the learned prototype.
Further, by learning the prototype and the transformation concurrently, the prototype adapts to the most dominant motion pattern, and the normalized data is aligned accordingly. 

\begin{figure*}[t!]
	\setlength{\tabcolsep}{0.5pt}  
	\begin{center}	
		\begin{tabular}{cc|ccccc}
			\hline
			\includegraphics[width=0.139\textwidth]{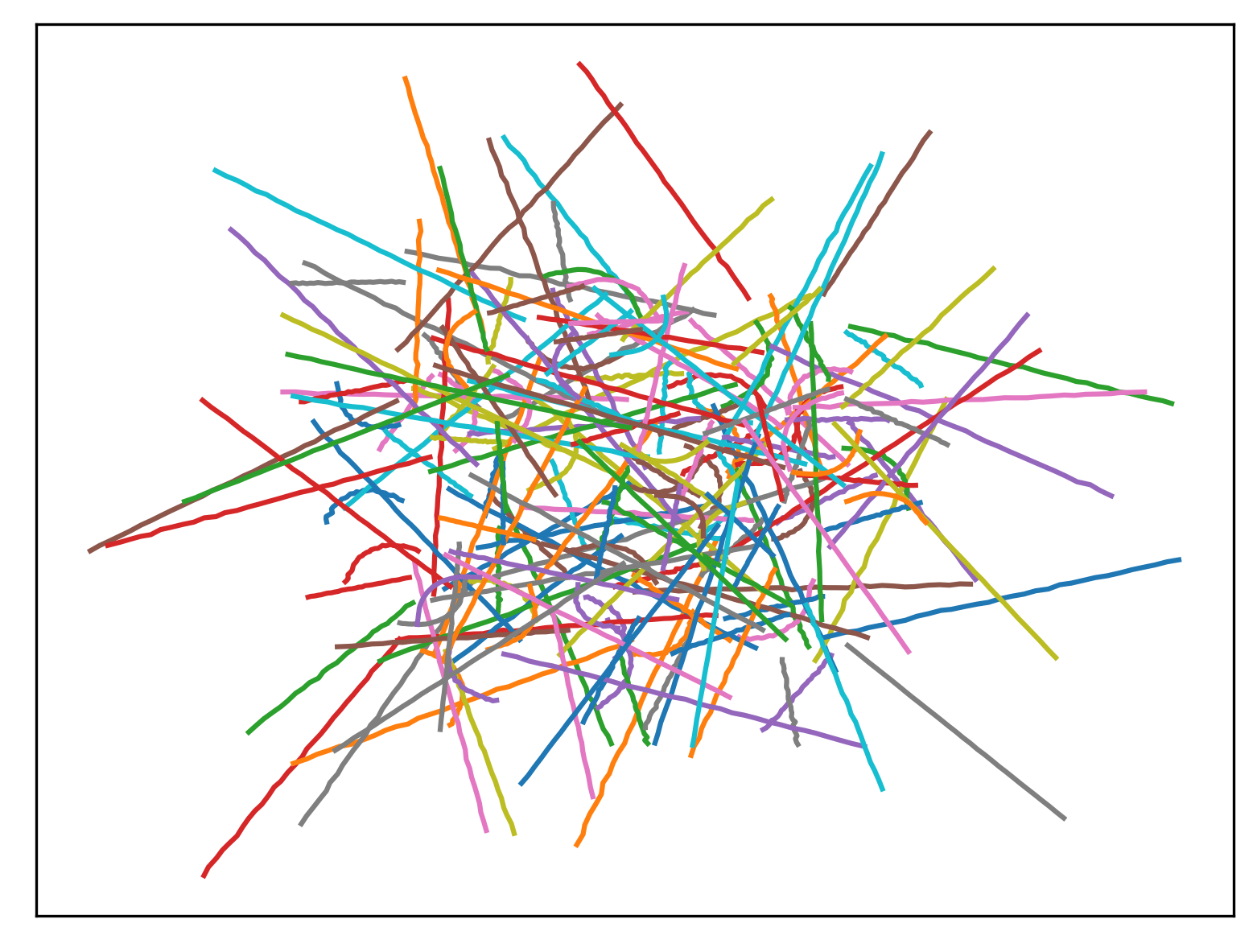} &
			\includegraphics[width=0.139\textwidth]{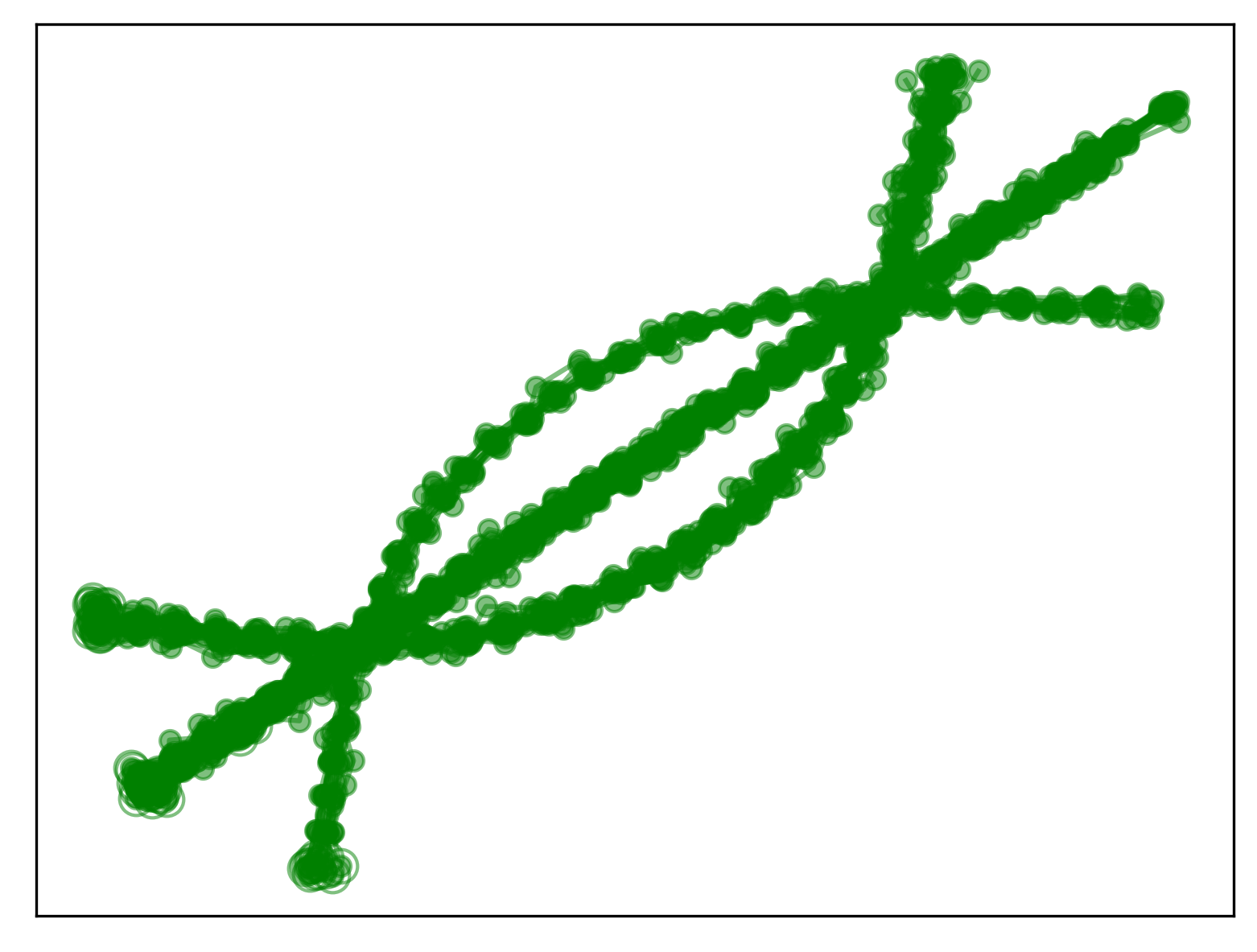} &
			\includegraphics[width=0.139\textwidth]{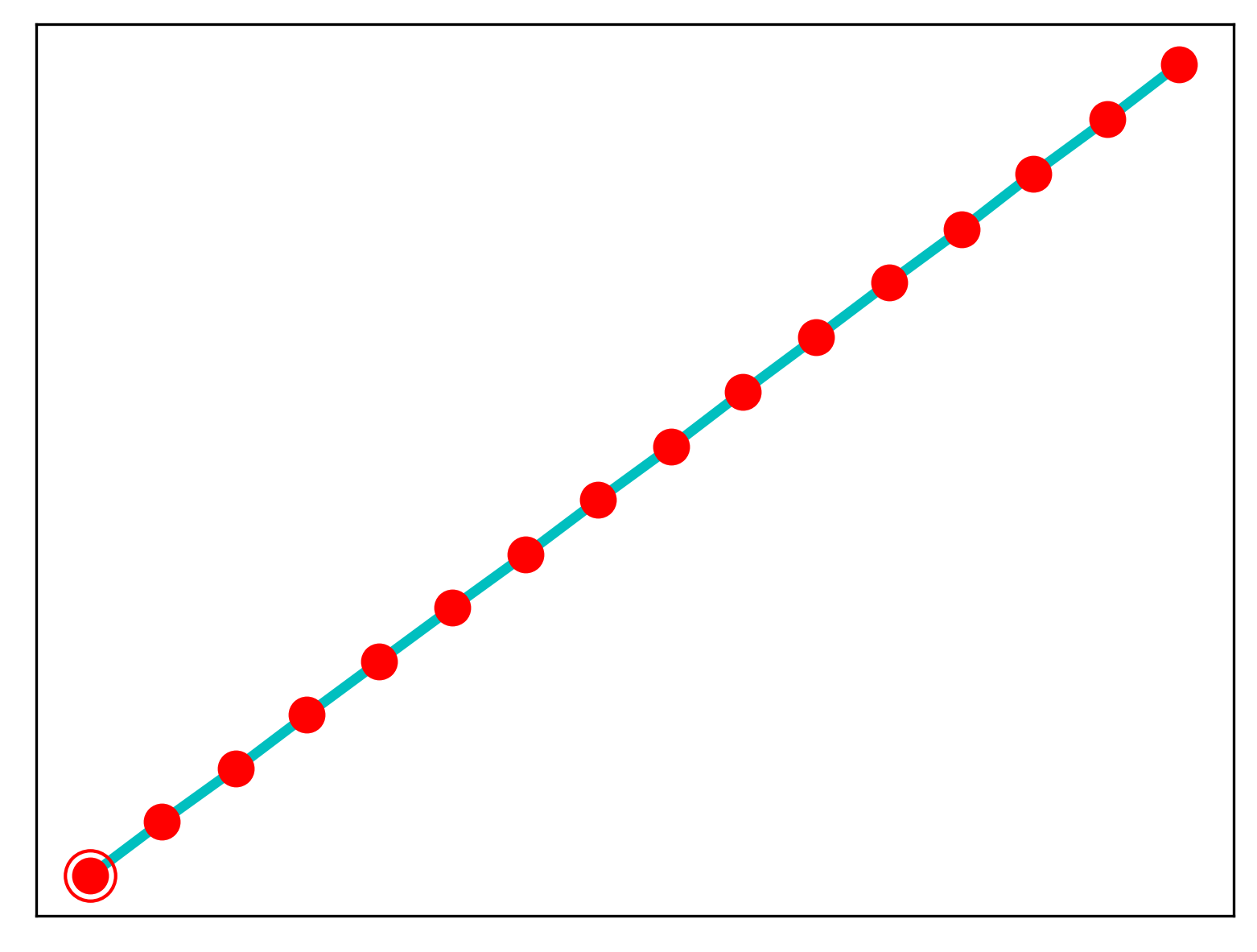} &
			\includegraphics[width=0.139\textwidth]{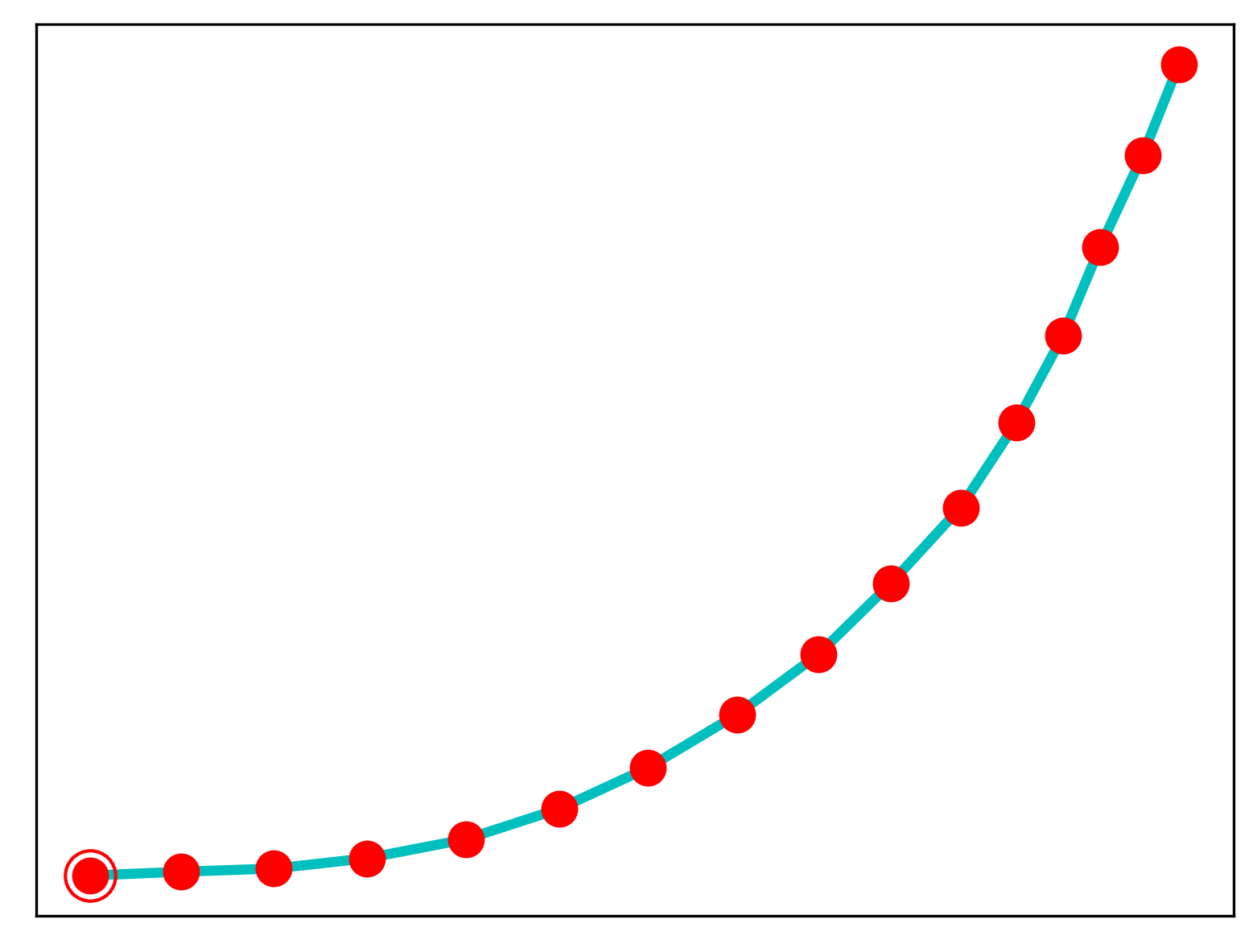} &
			\includegraphics[width=0.139\textwidth]{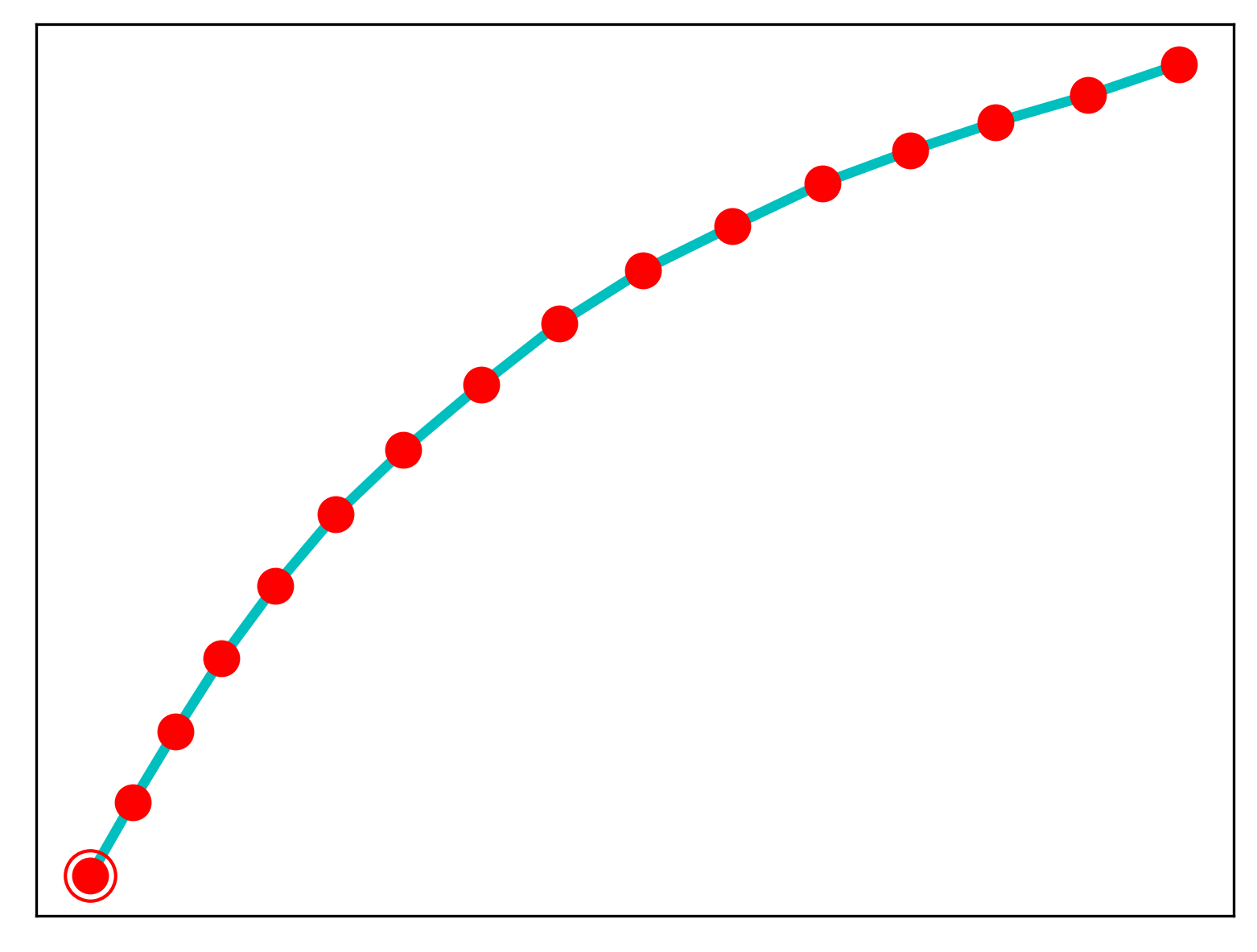} &
			\includegraphics[width=0.139\textwidth]{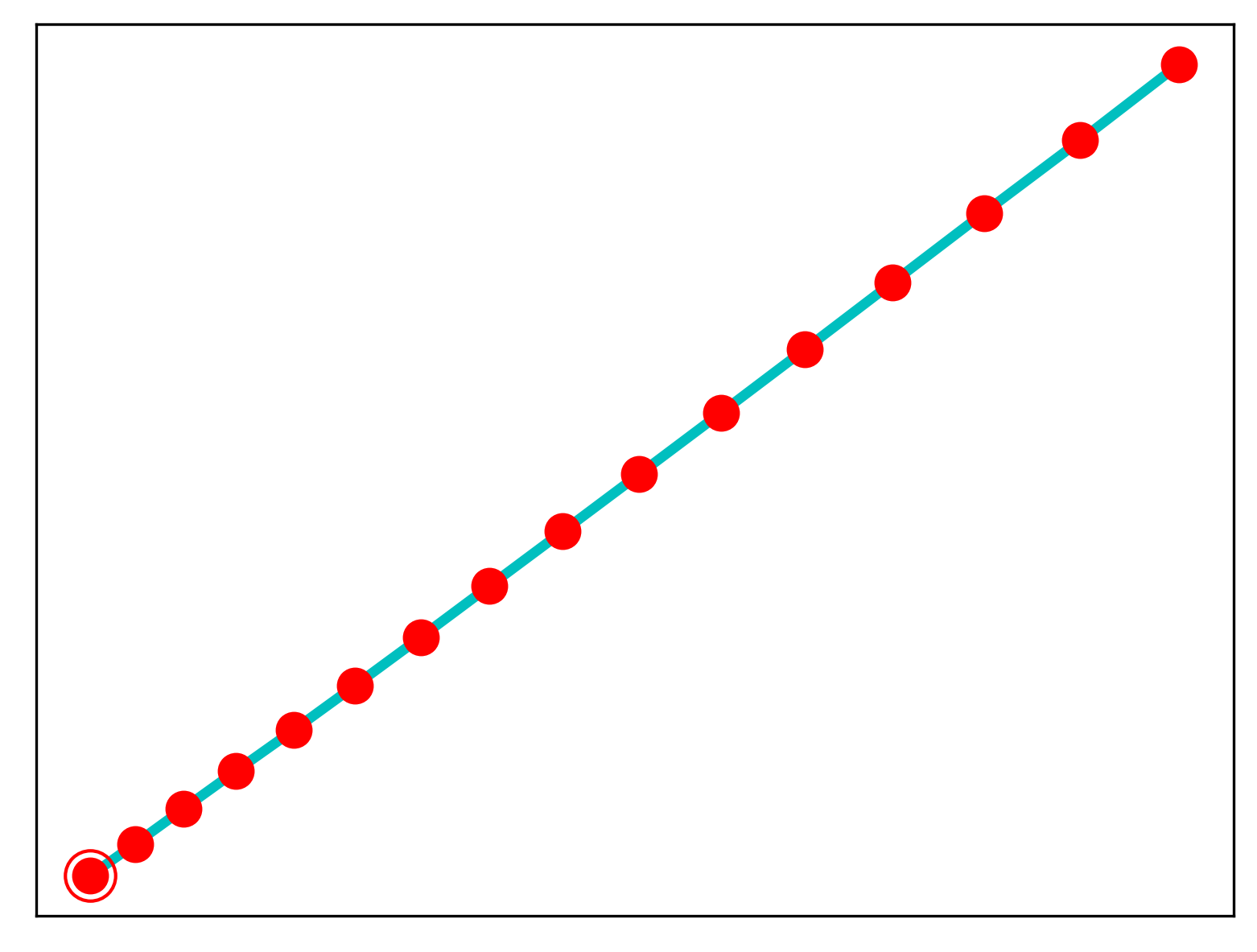} & \\
			
			\includegraphics[width=0.139\textwidth]{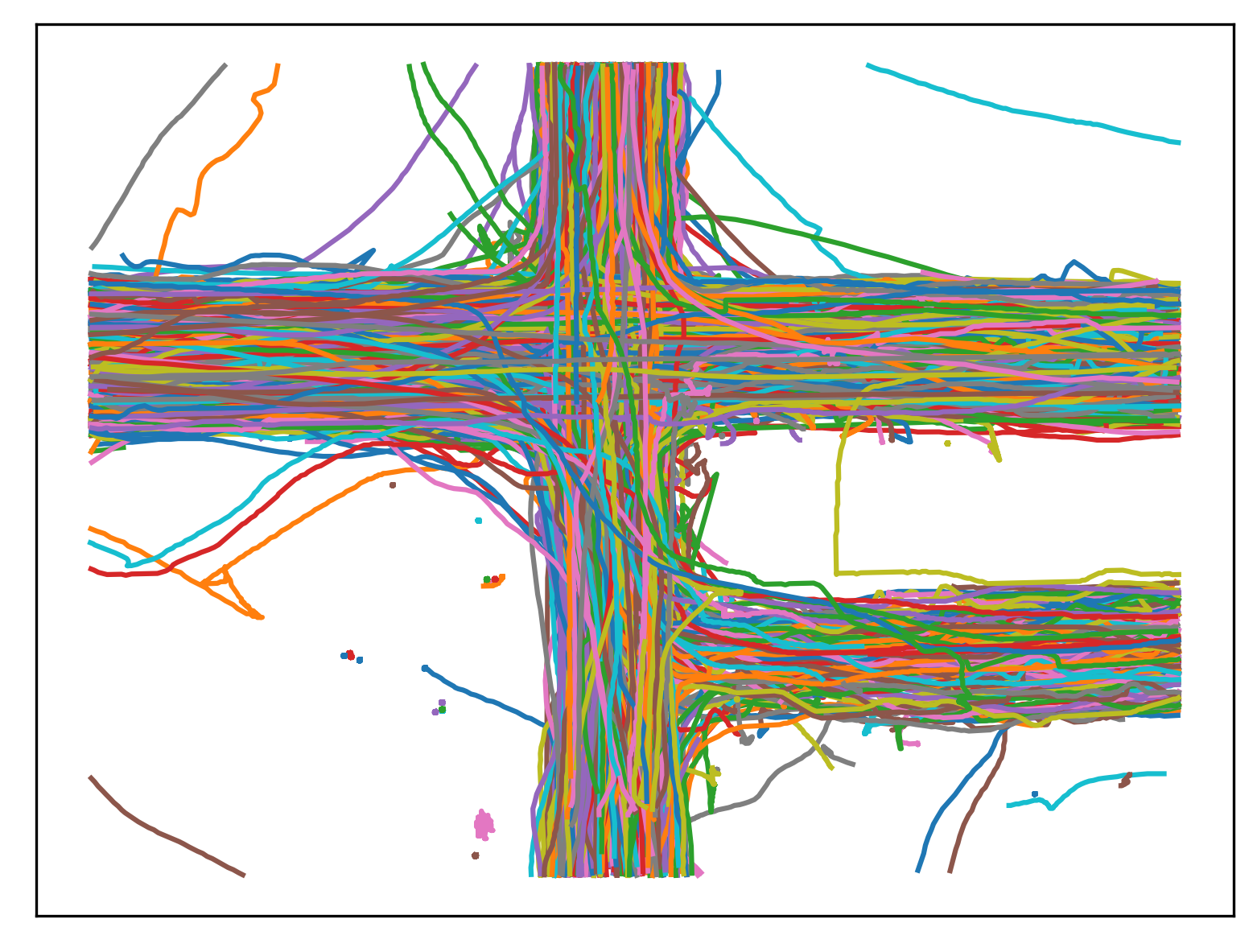} &
			\includegraphics[width=0.139\textwidth]{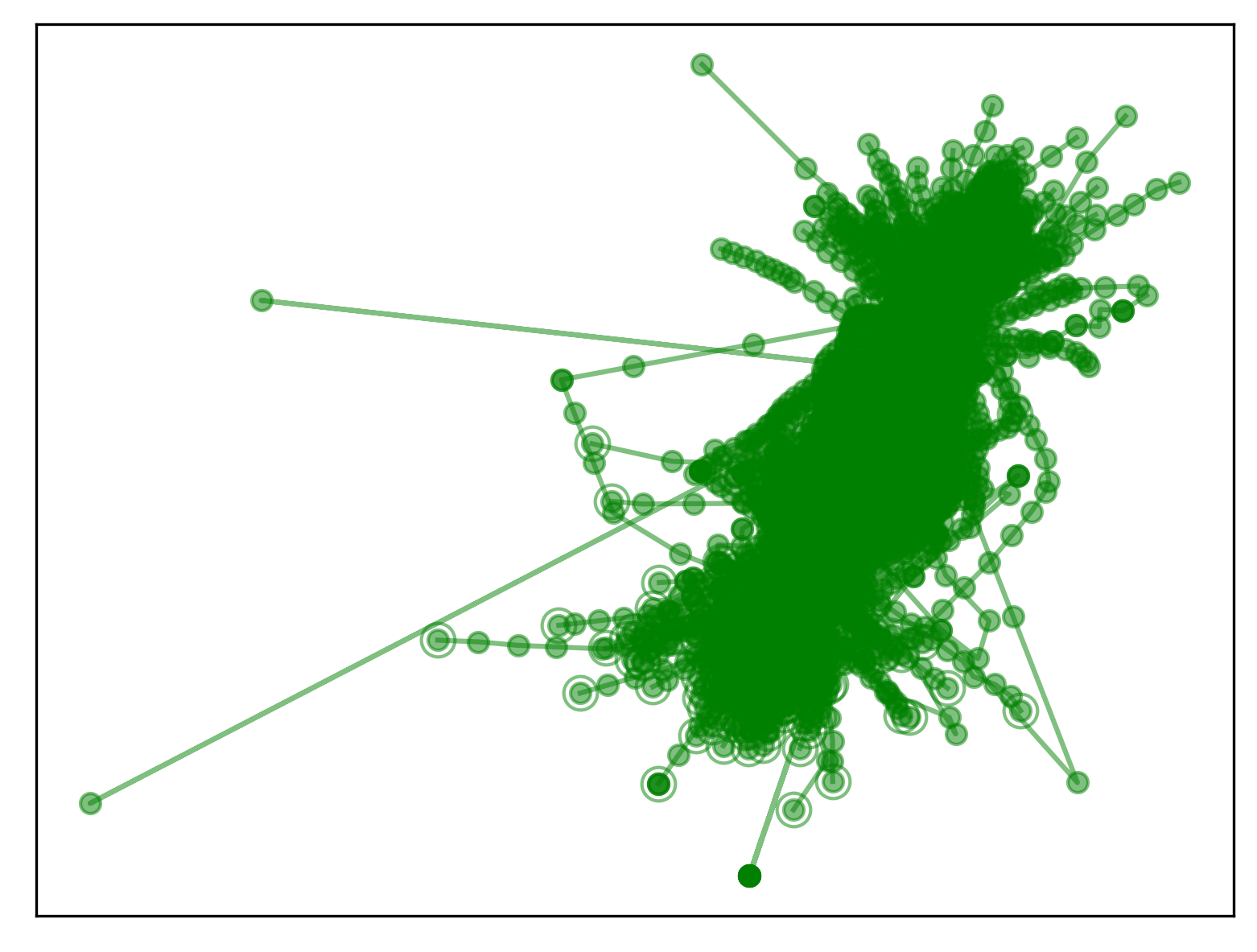} &	
			\includegraphics[width=0.139\textwidth]{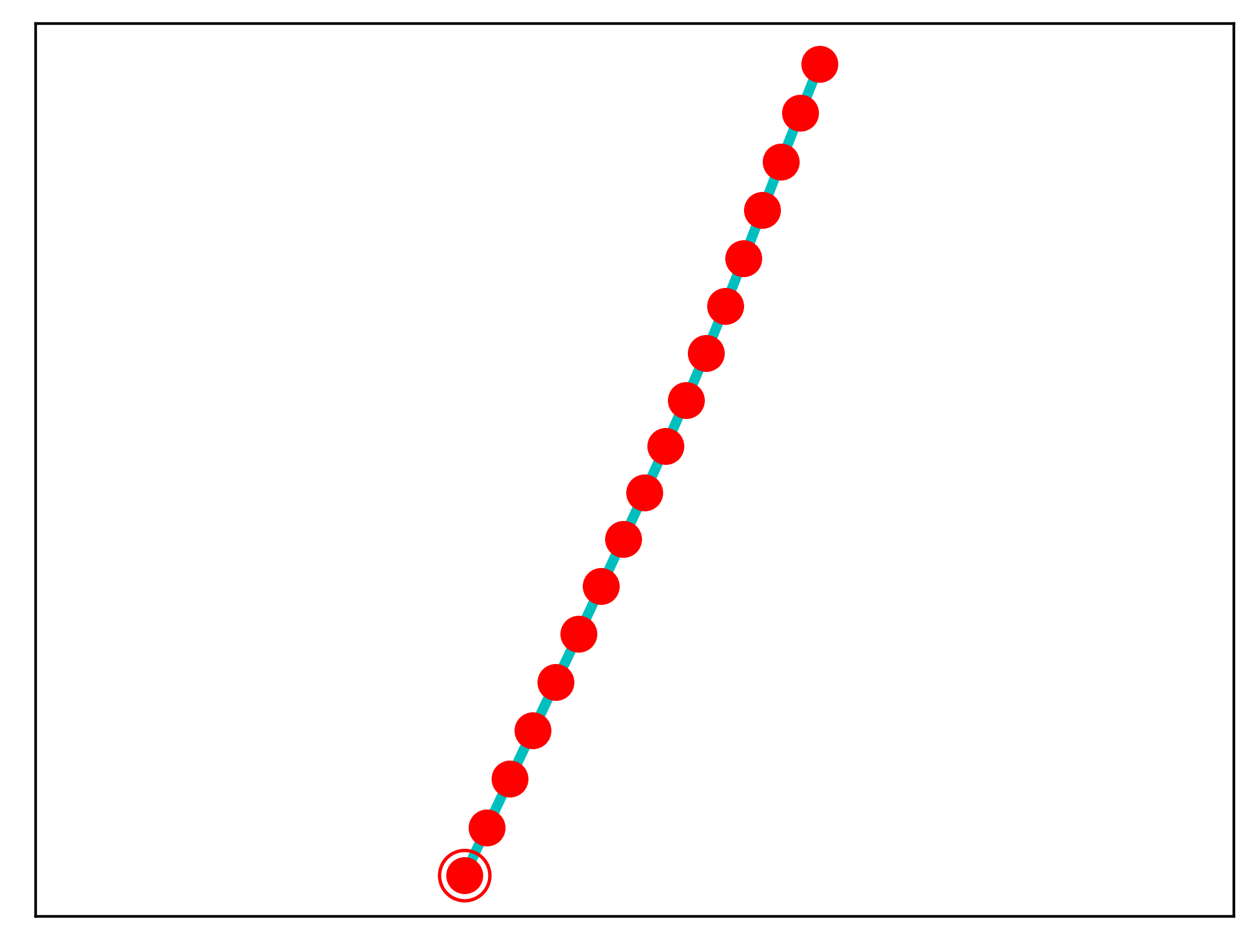} &
			\includegraphics[width=0.139\textwidth]{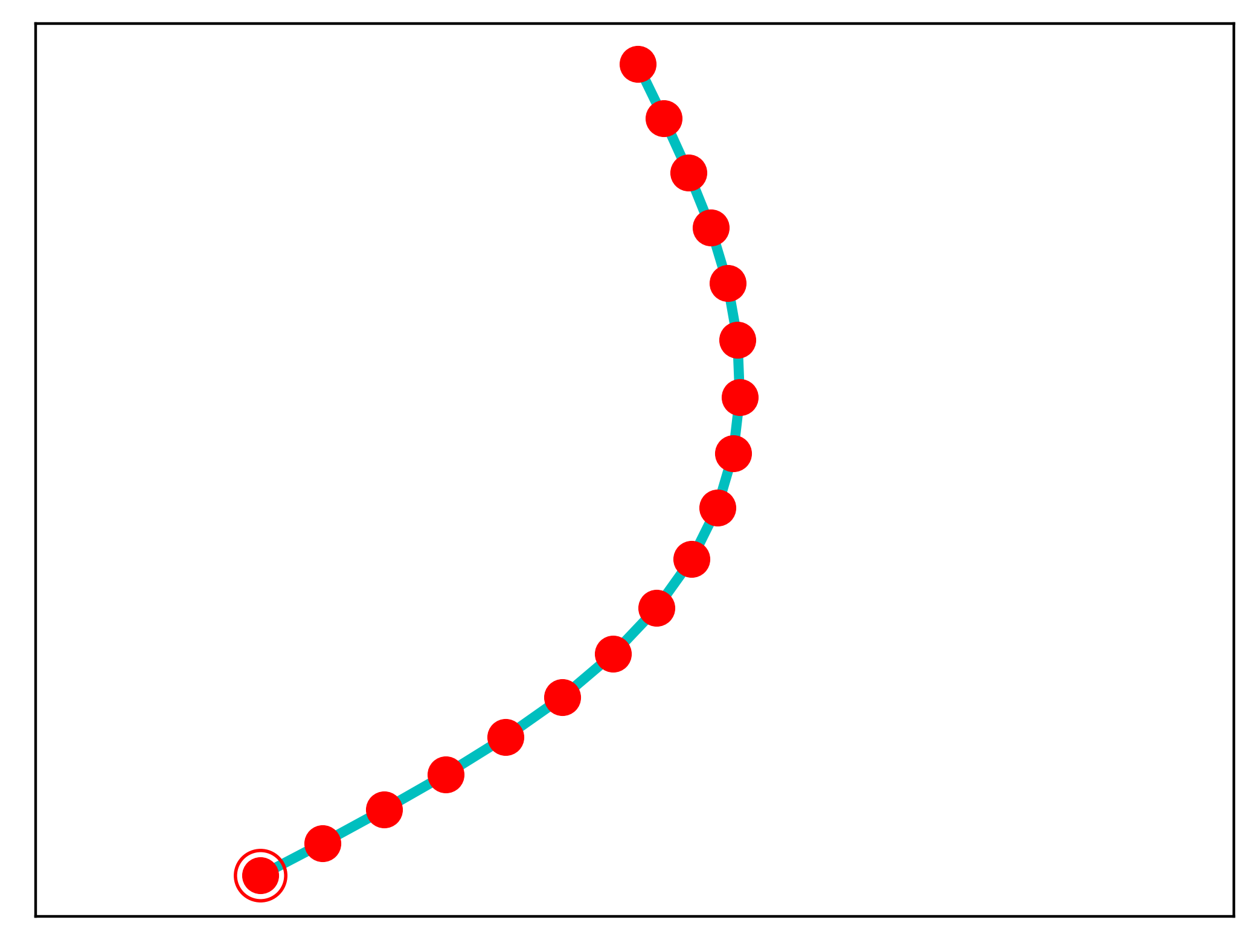} &
			\includegraphics[width=0.139\textwidth]{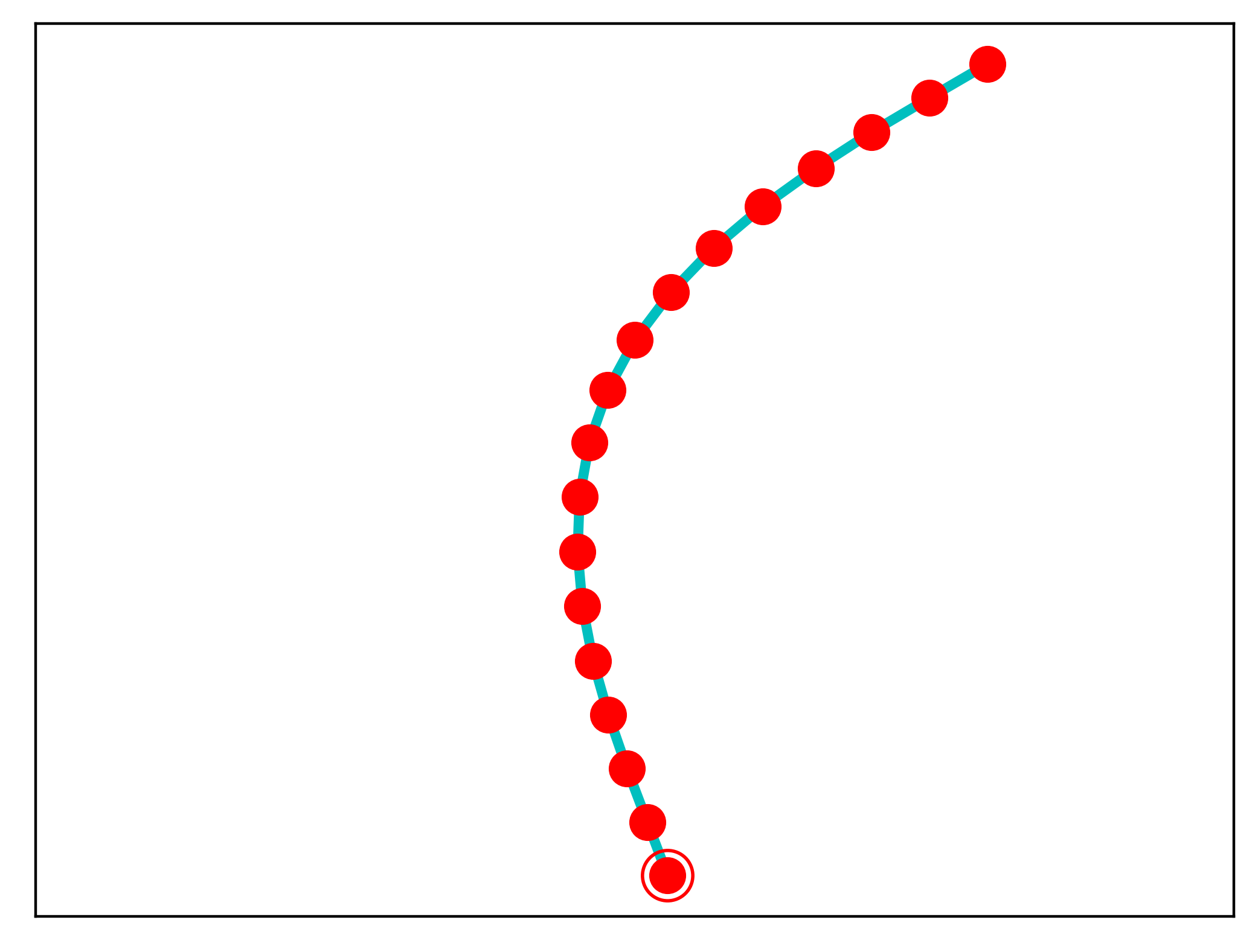} &
			\includegraphics[width=0.139\textwidth]{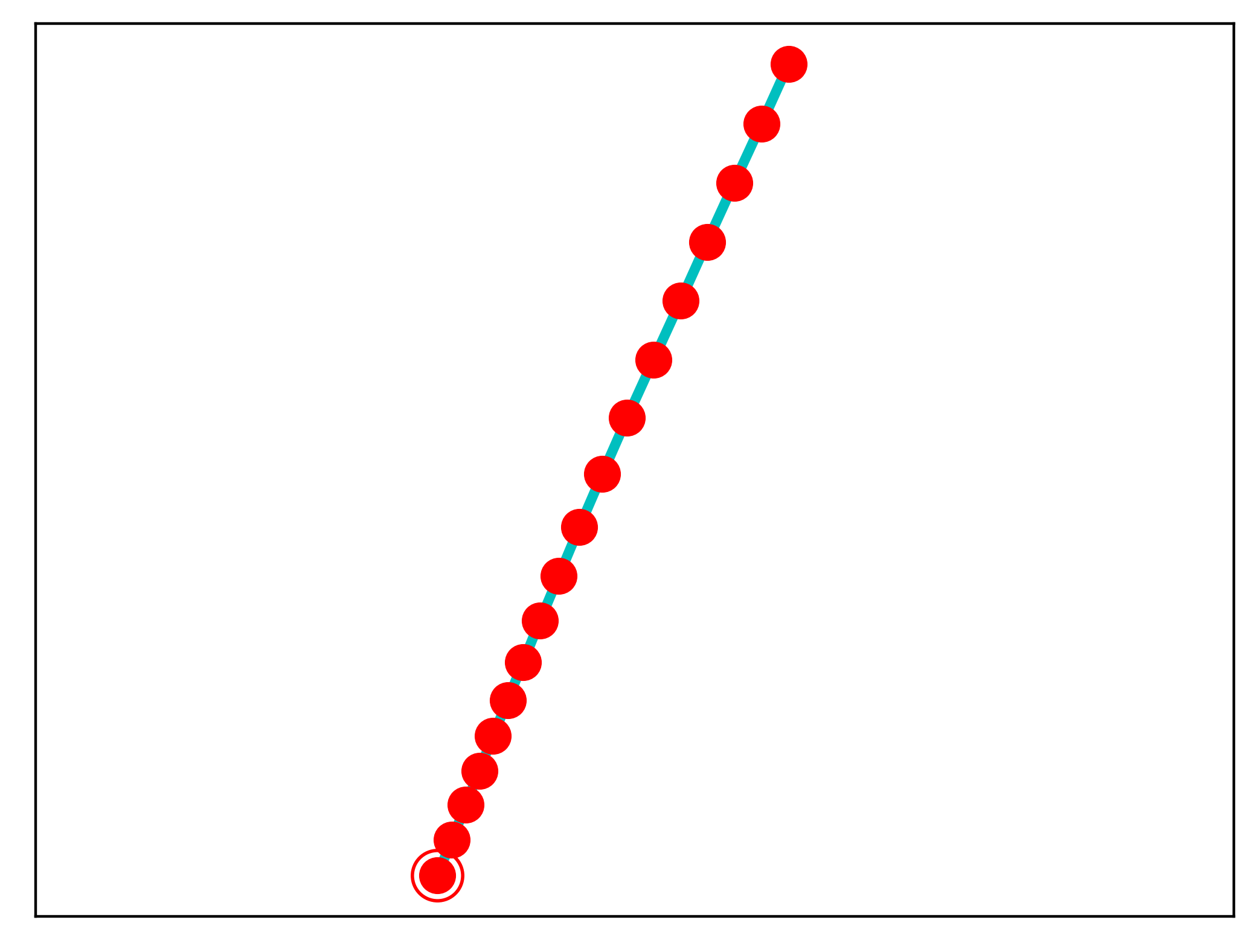} &
			\includegraphics[width=0.139\textwidth]{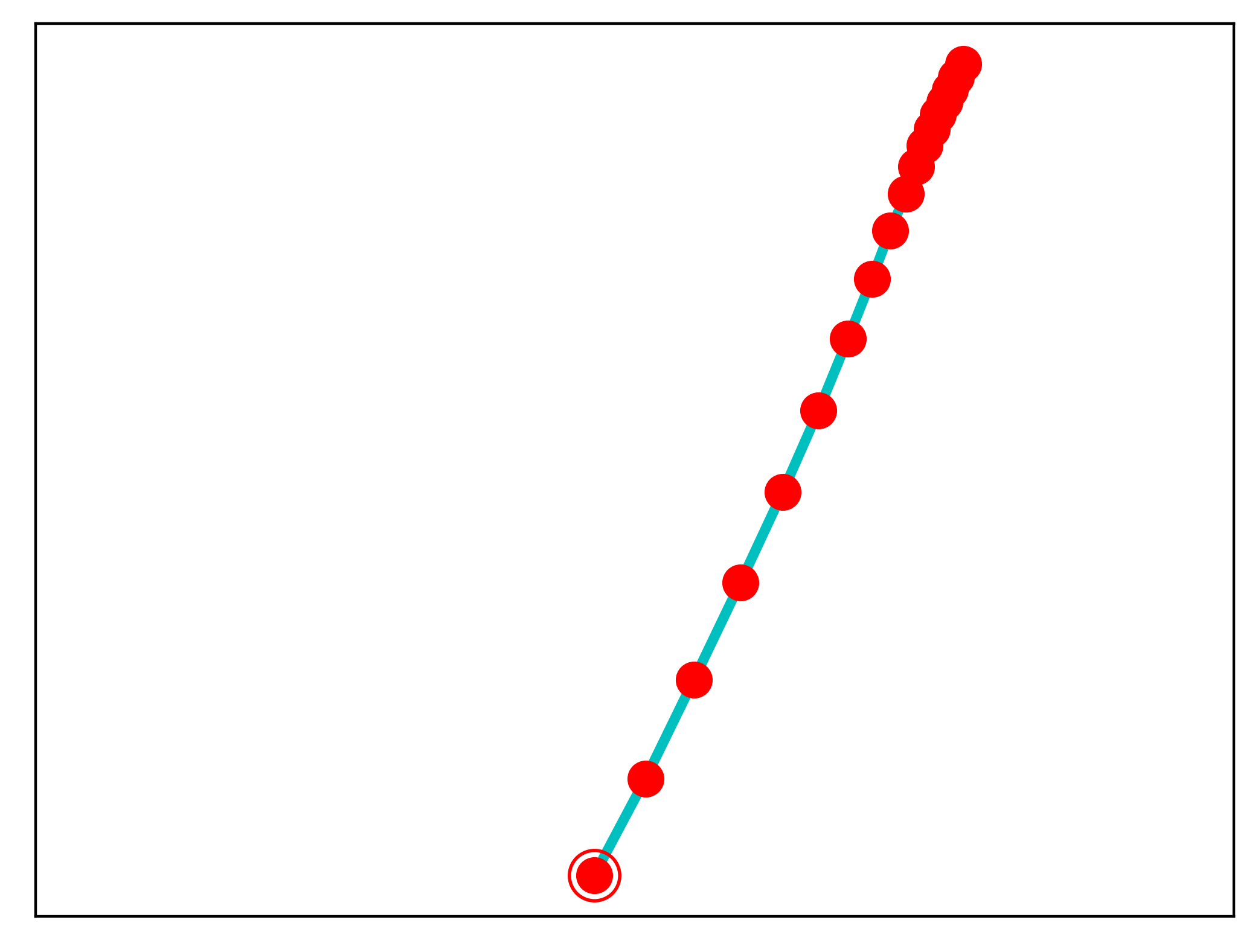}	\\
			\hline
			
			\small Data & \small Aligned Data & \small \small Constant & \small Curve (l) & \small Curve (r) & \small Acceleration & \small Deceleration \\
			&& \multicolumn{5}{c}{\small Prototypes}
		\end{tabular}
	\end{center}
	\caption{Data, aligned data and learned prototypes for the synthetic dataset SGTD (top) and the SDD-hyang dataset (bottom).
	For clarity only every 2nd or 3rd trajectory point is plotted for SGTD and SDD-hyang prototypes respectively.
	The sample ratios (left to right) for SGTD are $62.5\%$, $14.5\%$, $6.5\%$ and $16.5\%$, and for SDD-hyang $53\%$, $7.4\%$, $6.6\%$, $18.4\%$ and $14.6\%$.}
	\label{fig:results}
\end{figure*}

\section{Learning Vector Quantization} 
\label{sec:vq}
Clustering approaches can be applied after alignment, since the alignment approach distributes random errors homogeneously over the sequence. 
For clustering, a vector quantization (LVQ) approach is proposed, which is inspired by \cite{oord2017neural} and can directly be integrated in a deep learning framework.
Here, aligned samples are mapped onto $K$ prototypes $\mathcal{Z} = \{Z_1, ..., Z_K\}$, with $Z_k = \{\mathbf{z}^k_1, ..., \mathbf{z}^k_M\}$, in quantized space.
This results in a concise set of prototypes representing the given dataset.
The prototypes are learned by minimizing the mean squared error between the aligned samples and the respective closest prototype in quantized space:
\begin{align}
\begin{split}
	\mathcal{L}_{LVQ} &= \frac{1}{N \cdot M} \sum_{i=1}^{N}\sum^M_{j=1} \|\mathbf{x}^i_j - z_j(i)\|^2_2 + \gamma \cdot L_{global}, \\
	\text{with} \\
	z_j(i) &= \mathbf{z}^{argmin_k \frac{1}{M} \sum_{jj=1}^{M} \|\mathbf{x}^i_{jj} - \mathbf{z}^k_{jj}\|^2_2}_{j}, \\
	L_{global} &= \frac{1}{N \cdot M \cdot K} \sum_{i=1}^{N}\sum_{j=1}^{M}\sum_{k=1}^{K} \| \mathbf{x}^i_j - \mathbf{z}^k_j \|^2_2.
\end{split}
\end{align}
In addition, a regularization term $L_{global}$ is employed for driving all prototypes towards the global mean.
Intuitively, this moves low-support prototypes in more reasonable areas within quantized space, while the winner takes all loss in $\mathcal{L}_{LVQ}$ keeps them within range of relevant sample clusters.
Due to the fact that, it is unknown how many prototypes are necessary, unnecessary prototypes can be identified by their similarity to the dominant prototype and thus be sorted out later, in case $K$ is too large.

\textbf{Initialization.} Due to the winner takes all strategy, $\mathcal{L}_{LVQ}$ only updates prototypes that have supporting samples.
While the regularization term tries to account for this, initializing the $K$ prototypes appropriately helps to improve the quantization results.
Thus, a Forgy initialization is applied, i.e. randomly selected samples from the dataset are used as initial prototypes \cite{pena1999empirical}. 
This follows common practice of other vector quantization approaches, such as K-means \cite{lloyd1982kmeans}. 

\textbf{Refinement.} A heuristic refinement scheme is employed in order to remove unnecessary prototypes when $K$ was too large.
Starting with the prototype which is supported by most training samples, additional prototypes are added based on their dissimilarity to previously added prototypes and their impact on the global error given the refined set of prototypes.
By choosing prototypes that maximize the global error, a wider coverage of motion patterns is ensured.
Thus, unnecessary prototypes, especially those similar to the most dominant prototype, can be sorted out.

\section{Qualitative Evaluation}
\label{sec:eval}
As a first proof of concept, a qualitative evaluation is conducted on a synthetically generated dataset (\emph{SGTD}) and the combined \emph{hyang} scenes \cite{hug2017reliability} taken from the \emph{stanford drone dataset} (SDD, \cite{robicquet2016learning}).
SGTD consists of $200$ randomly distributed trajectories consisting of $M=31$ points\footnote{This number is arbitrary and does not affect the viability of the approach.}, with random orientations and different normally distributed speed profiles. 
It covers $4$ motion patterns: constant ($125$ samples), accelerated ($33$ samples), and curvilinear ($29$ (left) and $13$ (right) samples; direction change by $\alpha \sim \mathcal{U}(80, 90)$ degrees in either direction).
For SDD-hyang, the rate has been reduced to 6 samples per second.
This achieves a more noticeable difference between subsequent trajectory points.
Further, all trajectory segments containing $54$ points\footnote{Corresponds to the first quartile with respect to the number of trajectory points for all trajectories} are considered for training. 
Since this is a real-world dataset, the generating features are unknown.
Yet the occurrence of all basic motion patterns (constant, acceleration, decelerated and curvilinear motion) is expected reflecting its complexity. 
For both datasets, $K=10$ prototypes are used, providing sufficient capacity to capture all relevant features while also allowing to test the proposed regularization and refinement strategies.

The results of the alignment and the learned prototypes are depicted in figure \ref{fig:results}.
For SGTD (first row), it can be seen that all samples are aligned such that they follow a common mean orientation and similar samples are pooled together.
All 4 prototypes necessary for modeling the dataset were found with correct association between data and prototypes.
For SDD-hyang, there is not an as clear separation between different motion patterns. 
However, a similar behavior can be observed when looking at the aligned data.
Further, as expected, the quantization yields an additional prototype corresponding to a decelerated motion.

\section{Discussion and Future Work}
At last, given the results in section \ref{sec:eval}, future work building on the proposed approach is discussed.

\textbf{Quantifying Dataset Complexity.}
A thorough complexity analysis of standard benchmarking datasets for trajectory prediction, deriving relevant complexity factors from the proposed prototype representation should be considered.
These factors could involve for example the variety of prototypes or the sample association bias.
Analyzing the complexity of these datasets could offer more insight into their viability for evaluating certain aspects of machine learning models.

\textbf{Deriving a Benchmark from Insights.}
Insights gained from dataset complexity analysis could be used for deriving a new benchmark for trajectory prediction, that specifically targets different aspects of a given model.
This could provide a viable complementary benchmark to the recently published \emph{TrajNet++} benchmark (mentioned in \cite{rudenko2019human}; successor of \emph{TrajNet} \cite{sadeghiankosaraju2018trajnet}).

\textbf{Data Preprocessing.}
The proposed approach may also serve as universal preprocessing stage for trajectory prediction models.
Thus an experimental study should be conducted for comparing preprocessing approaches.

\addtolength{\textheight}{-12cm}   


\bibliographystyle{IEEEtran}
\bibliography{IEEEabrv,bibliography}

\begin{thebibliography}{1}
\providecommand{\url}[1]{#1}
\csname url@rmstyle\endcsname
\providecommand{\newblock}{\relax}
\providecommand{\bibinfo}[2]{#2}
\providecommand\BIBentrySTDinterwordspacing{\spaceskip=0pt\relax}
\providecommand\BIBentryALTinterwordstretchfactor{4}
\providecommand\BIBentryALTinterwordspacing{\spaceskip=\fontdimen2\font plus
\BIBentryALTinterwordstretchfactor\fontdimen3\font minus
  \fontdimen4\font\relax}
\providecommand\BIBforeignlanguage[2]{{%
\expandafter\ifx\csname l@#1\endcsname\relax
\typeout{** WARNING: IEEEtran.bst: No hyphenation pattern has been}%
\typeout{** loaded for the language `#1'. Using the pattern for}%
\typeout{** the default language instead.}%
\else
\language=\csname l@#1\endcsname
\fi
#2}}

\bibitem{rudenko2019human}
A.~Rudenko, L.~Palmieri, M.~Herman, K.~M. Kitani, D.~M. Gavrila, and K.~O.
  Arras, ``Human motion trajectory prediction: A survey,'' \emph{arXiv preprint
  arXiv:1905.06113}, 2019.

\bibitem{oord2017neural}
A.~van~den Oord, O.~Vinyals, \emph{et~al.}, ``Neural discrete representation
  learning,'' in \emph{Advances in Neural Information Processing Systems},
  2017, pp. 6306--6315.

\bibitem{pena1999empirical}
J.~M. Pena, J.~A. Lozano, and P.~Larranaga, ``An empirical comparison of four
  initialization methods for the k-means algorithm,'' \emph{Pattern recognition
  letters}, vol.~20, no.~10, pp. 1027--1040, 1999.

\bibitem{lloyd1982kmeans}
S.~Lloyd, ``Least squares quantization in pcm,'' \emph{IEEE transactions on
  information theory}, vol.~28, no.~2, pp. 129--137, 1982.

\bibitem{hug2017reliability}
R.~Hug, S.~Becker, W.~H{\"u}bner, and M.~Arens, ``On the reliability of
  lstm-mdl models for pedestrian trajectory prediction,'' \emph{VIIth
  International Workshop on Representation, analysis and recognition of shape
  and motion FroM Image data (RFMI 2017)}, 2017.

\bibitem{robicquet2016learning}
A.~Robicquet, A.~Sadeghian, A.~Alahi, and S.~Savarese, ``Learning social
  etiquette: Human trajectory understanding in crowded scenes,'' in
  \emph{European conference on computer vision}.\hskip 1em plus 0.5em minus
  0.4em\relax Springer, 2016, pp. 549--565.

\bibitem{sadeghiankosaraju2018trajnet}
A.~Sadeghian, V.~Kosaraju, A.~Gupta, S.~Savarese, and A.~Alahi, ``Trajnet:
  Towards a benchmark for human trajectory prediction,'' \emph{arXiv preprint},
  2018.

\end{thebibliography}

\end{document}